\documentclass[letterpaper, 10 pt, conference]{ieeeconf}  % Comment this line out if you need a4paper

\IEEEoverridecommandlockouts                              % This command is only needed if 
\usepackage{graphicx} % Required for including pictures
\usepackage{float}    % For tables and other floats
\usepackage{verbatim} % For comments and other
\usepackage{amsmath}  % For math
\usepackage{amssymb}  % For more math
\usepackage[font=footnotesize]{caption}
\usepackage[ruled,vlined,linesnumbered]{algorithm2e}
\usepackage{multirow}
\usepackage{array}
\usepackage{booktabs}
\usepackage{hyperref}
\usepackage[dvipsnames]{xcolor}
\usepackage{dsfont}
\usepackage{siunitx}
\usepackage{tensor}
\usepackage{mathtools}

\title{\LARGE \bf
Learning Model Predictive Control with Error Dynamics Regression for Autonomous Racing
}

\author{Haoru Xue$^{1}$, Edward L. Zhu$^{2}$, John M. Dolan$^{1}$, and Francesco Borrelli$^{2}$% <-this % stops a space
\thanks{$^{1}$Haoru Xue and John M. Dolan are with the Robotics Institute, Carnegie Mellon University, Pittsburgh, PA, 15213, USA
        {\{\tt\small haorux, jdolan\}@andrew.cmu.edu}}%
\thanks{$^{2}$Edward L. Zhu and Francesco Borrelli are with the Department of Mechanical Engineering, University of California Berkeley,
        Berkeley, CA 94720, USA
        {\{\tt\small edward.zhu, fborrelli\}@berkeley.edu}}%
}

\begin{document}

\maketitle
\thispagestyle{empty}
\pagestyle{empty}

%%%%%%%%%%%%%%%%%%%%%%%%%%%%%%%%%%%%%%%%%%%%%%%%%%%%%%%%%%%%%%%%%%%%%%%%%%%%%%%%
\begin{abstract}
This work presents a novel Learning Model Predictive Control (LMPC) strategy for autonomous racing at the handling limit that can iteratively explore and learn unknown dynamics in high-speed operational domains. We start from existing LMPC formulations and  modify the system dynamics learning method. In particular, our approach uses a nominal, global, nonlinear, physics-based model with a local, linear, data-driven learning of the error dynamics. We conducted experiments in simulation and on 1/10th scale hardware, and deployed the proposed LMPC on a full-scale autonomous race car used in the Indy Autonomous Challenge (IAC) with closed loop experiments at the Putnam Park Road Course in Indiana, USA.
The results show that the proposed control policy  exhibits improved robustness to parameter tuning and data scarcity. Incremental and safety-aware exploration toward the limit of
handling and iterative learning of the vehicle dynamics in high-speed domains is observed both in simulations and experiments.
\end{abstract}

\section{Introduction}

Recent progress in autonomous racing research has empowered full-size autonomous race cars at or near the top speed of a professional human racing series. Algorithms that were previously experimented with on small-scale vehicles \cite{okelly_f1tenth_2020} are deployed and further researched on professional race car platforms. Specifically, the teams in the Indy Autonomous Challenge (IAC) \cite{noauthor_indy_2023}, starting in 2021, have been demonstrating multi-agent competition above 240 \si[per-mode=symbol]{\kilo\meter\per\hour} \cite{wischnewski2022indy}.

In the field of control, which is tasked with trajectory tracking at the limit of vehicle dynamics handling, various Model Predictive Control (MPC) techniques remain the state-of-the-art on full-size race car systems. These variations of MPC algorithms solve a finite-horizon optimal control problem (FHOCP) by computing optimal actuation commands for a vehicle modeled with kinematic or dynamic models, subject to certain track boundary and state constraints \cite{wischnewski_tube-mpc_2023,kabzan_amz_2020,betz_tum_2023,funke_up_2012}. However, for high-speed racing, the performance of classic, nonadaptive MPC methods is ultimately bounded by the fidelity of the vehicle dynamics model. Inaccurate vehicle models will not only induce suboptimal behavior such as weaving, hunting, and steady-state tracking error, but also become dangerous at the handling limit, especially when the model overpredicts this limit. This introduces a chicken-and-egg problem for autonomous racing research at the handling limit: to safely achieve higher speed levels requires more vehicle dynamics data; but to acquire high-speed data requires the controller to achieve higher speed levels in the first place. Therefore, the motivation of this work is to break the circular problem mentioned earlier by performing incremental and safety-aware exploration toward the limit of handling and iterative learning of the vehicle dynamics in high-speed domains.

\begin{figure}[t]
     \centering
     % \vspace{-1.cm}
     {\includegraphics[trim={0 0 0 0cm},clip,width=\columnwidth]{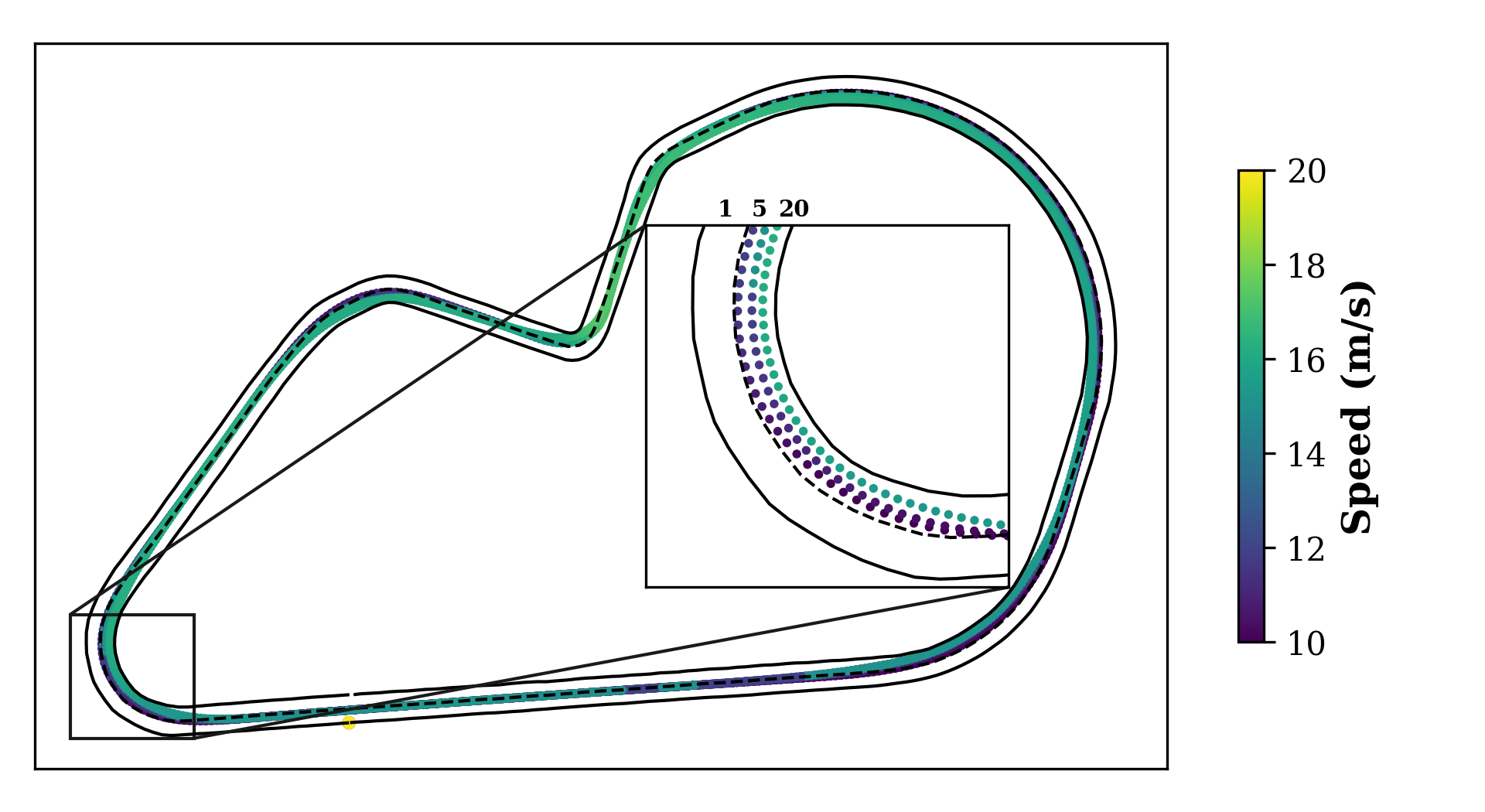}}
     % \vspace{-1.3cm}
     \caption{Overlay of the IAC car's trajectory running LMPC with error dynamics regression. The data are collected at the Putnam Park Road Course in Indiana, USA. The zoomed-in image of the hairpin turn shows iterative improvement towards optimal cornering in the 1st, 5th, and 20th lap.}
     \label{fig:iac_vis}
     \vspace{-0.5cm}
\end{figure}

In this work, we present a modification to the LMPC approach presented in \cite{rosolia_learning_2020} where we improve significantly upon the model learning strategy. In particular, we propose to learn the error dynamics between a given nominal model and the state evolution of the true system collected during run-time. This is in contrast with the prior approach, which attempts to learn a model of the true system directly. We show through simulation and hardware experiments on a 1/10th scale vehicle that our approach exhibits improved robustness to parameter tuning and reduced modeling errors in scenarios when data are scarce for model learning. We finally demonstrate the effectiveness of our approach on a full-size IAC autonomous race car, the results of which are shown in Figure~\ref{fig:iac_vis}. An open source C++ implementation of our work has been made available at \url{https://github.com/MPC-Berkeley/Racing-LMPC-ROS2}.

% \subsection{Previous Works}

\noindent \emph{Related Work:} Learning unknown or changing system dynamics has been studied within some classic control frameworks such as adaptive control \cite{bujarbaruah2018adaptive,sasfi2023robust}. System identification is usually part of the controller design, which reasons about some modeling error or noise in the system and adapts the system model to account for them. In autonomous racing, previous works have proposed dynamics learning strategies using Gaussian Processes \cite{kabzan2019learning,hewing_cautious_2018,ning2023vehicle} or implicitly with Gaussian noise assumption \cite{kalaria2022delay}. The method is further extended to model multi-vehicle interactions and solve for the optimal overtaking maneuver \cite{zhu2023gaussian,brüdigam2021gaussian}.

Another branch of works exploits the repetitive nature of autonomous racing in a single-vehicle scenario, and improves controller performance without any prior assumption about the distribution of the modeling error or noise. Data-driven methods, such as iterative learning control, can optimize the tracking of a repetitive trajectory given data from previous iterations by solving for control variable corrections \cite{bristow2006ilc}. More recently, the learning model predictive control (LMPC) framework has been proposed with a direct regression of local affine dynamics \cite{rosolia_learning_2017,brunner2017repetitive,rosolia_learning_2020}. This combination of LMPC and regressive dynamics learning can iteratively learn the unknown system dynamics and reduce lap-time without an explicit prior reference trajectory or a nominal dynamics model. The approach has also been demonstrated on scaled vehicle hardware platforms in \cite{rosolia_learning_2017}. 

\section{Problem Formulation}

In this work, we model the vehicle using the planar dynamic bicycle model \cite{Kong2015,christ_time-optimal_2021} with state and input vectors $x = [v_x, v_y, \omega_z, e_{\psi}, s, e_y]^{\top}$, $u = [a, \delta]^\top$,
% \begin{align*}
%     x &= [v_x, v_y, \omega_z, e_{\psi}, s, e_y]^{\top} \in \mathbb{R}^6 \\
%     u &= [a, \delta]^\top \in \mathbb{R}^2
% \end{align*}
where $v_x$, $v_y$, and $\omega_z$ are the vehicle's longitudinal velocity, lateral velocity, and yaw rate at the center of gravity. $s$, $e_y$, and $e_\psi$ are the pose of the vehicle expressed in the Frenet frame w.r.t. a parametric path $\tau:[0, L] \mapsto \mathbb{R}^2$, which is assumed to be continuously twice differentiable and of total length $L$. In particular, when given an arclength $s \in [0, L]$, $\tau$ returns the global $x$-$y$ position of the path. In the case where $\tau$ forms a closed circuit, i.e., $\tau(0) = \tau(L)$, $\tau'(0) = \tau'(L)$, $\tau''(0) = \tau''(L)$, where $\tau'$ and $\tau''$ are the element-wise first and second derivatives of $\tau$, we define the circuit as $\bar{\tau}(s) = \tau(\text{mod}(s, L))$ for $s \geq 0$, where mod is the modulo operator. Given a path $\tau$ and a vehicle with global position $p = (x,y)$ and heading $\psi$, 
the Frenet frame pose is defined as the vehicle's path progress $s(p) = \arg \min_s \|\tau(s) - p\|_2$, its lateral deviation from the path $e_y(p) = \min_s \|\tau(s) - p\|_2$, and its heading deviation from the path tangent $e_\psi(p,\psi) = \psi - \arctan(\tau'(s(p)))$.
The inputs to the model are the longitudinal acceleration $a$ and the front steering angle $\delta$. We write the discretized nonlinear dynamics as:
\begin{align} \label{eq:nominal_dynamics_model}
    x_{k+1} = f(x_k, u_k)
\end{align}
The vehicle's state and input are subject to the constraints $\mathcal{X} = \{x \ | \ -W/2 \leq e_y \leq W/2\}$ and $\mathcal{U} = \{u \ | \ u_l \leq u \leq u_u\}$, 
which describe the boundary constraints for a track of constant width $W$ and the limits on achievable acceleration and steering angle.

The objective of this work is to design a control policy which solves the infinite-horizon, minimium lap time, optimal control problem:
\begin{subequations} \label{eq:inf_horizon_min_lap_prob}
    \begin{align}
        T = \min_{u_0, u_1, \dots} \ & \ \sum_{k=0}^{\infty} \mathds{1}_{\mathcal{F}}(x_k) \label{eq:inf_horizon_min_lap_prob_cost} \\
        \text{subject to} \ & \ x_0 = \bar{x}, \\
        & \ x_{k+1} = f(x_k, u_k), \label{eq:inf_horizon_min_lap_prob_dyn}\\
        & \ x_k \in \mathcal{X}, \ u_k \in \mathcal{U}, \label{eq:inf_horizon_min_lap_prob_constr}
    \end{align}
\end{subequations}
where $\bar{x} \in \mathcal{S} = \{x \ | \ s = 0\}$ is a state at the start of the track and $\mathcal{F} = \{ x \ | \ s \geq L\}$ is the set of states beyond the finish line of the circuit. If for some time step $k$, $x_k \in \mathcal{F}$ for the first time, then the vehicle has finished the lap and has transitioned onto the next lap. $\mathds{1}_{\mathcal{F}}$ is the indicator function for the set $\mathcal{F}$, which is defined as
\begin{align*}
    \mathds{1}_{\mathcal{F}}(x) = 
    \begin{cases}
        0 \quad \text{if} \ x \in \mathcal{F} \\
        1 \quad \text{otherwise}
    \end{cases}
\end{align*}

Due to the infinite horizon cost in \eqref{eq:inf_horizon_min_lap_prob_cost}, it should be clear that the optimal control problem posed in \eqref{eq:inf_horizon_min_lap_prob} is not amenable to online implementation. Moreover, it is likely that there is mismatch between the vehicle model in \eqref{eq:inf_horizon_min_lap_prob_dyn} and the true vehicle dynamics, the effects of which would be especially apparent at high speeds, where the vehicle is operating at the limit of handling. To address these challenges, previous work on LMPC \cite{brunner2017repetitive, rosolia2017learning} proposed a finite-horizon convex approximation of \eqref{eq:inf_horizon_min_lap_prob}, which uses state and input trajectories from prior laps to synthesize a convex terminal cost-to-go function and target set. The same data are additionally used to identify affine time-varying (ATV) models which approximate the true dynamics of the vehicle. In this work, we carry over the same approach for terminal cost-to-go function and target set synthesis, but propose a modification to the system identification procedure which results in an LMPC control policy that achieves similar performance to \cite{brunner2017repetitive, rosolia2017learning} w.r.t. minimum lap time, while exhibiting a significant improvement in robustness to tuning parameters.

\section{Local Learning Model Predictive Control}

In this section, we briefly describe the procedure for synthesizing LMPC policies over iterations (or laps) of the minimum time control task. For further details, we refer the reader to  \cite{brunner2017repetitive,rosolia2017learning}. 

The main idea of LMPC is to use the data from iterations $1$ to $j-1$ to synthesize a finite-horizon local convex approximation of \eqref{eq:inf_horizon_min_lap_prob} for iteration $j$, which is then solved in a receding horizon manner. Let us first denote the available dataset at iteration $j$ as $\mathcal{D}^j = (\mathbf{x}^j, \mathbf{u}^j) \cup \mathcal{D}^{j-1}$, where $\mathbf{x}^j = \{x_0^j, x_1^j, \dots, x_{T^j}^j\}$ and $\mathbf{u}^j = \{u_0^j, u_1^j, \dots, u_{T^j}^j\}$ are the closed-loop state and input sequence for iteration $j$ and $T^j$ denotes the time step at which the lap was completed for iteration $j$, i.e., the first time step where $\mathds{1}_{\mathcal{F}}(x_k) = 0$. Note that on the first iteration, the dataset $\mathcal{D}^0$ may be initialized using human-driven laps or closed-loop trajectories from a simple low-speed center-line tracking controller. In addition, we assume that the trajectories stored in $\mathcal{D}^j$ are feasible w.r.t. the constraints \eqref{eq:inf_horizon_min_lap_prob_constr} and reach the finish line in a finite amount of time.

\subsection{Local Convex Target Set}

As the closed-loop trajectories stored in $\mathcal{D}^j$ are from the actual vehicle, it is straightforward to see that if we can control the vehicle to any state in $\mathcal{D}^j$, then there exists a known feasible control sequence which can be used to complete the lap from that state. LMPC uses this fact to construct a safe set as a discrete collection of the states in $\mathcal{D}^j$, which is then used as the terminal set for a FHOCP \cite{rosolia2017learning}. In this work, we construct convex terminal sets which are local about a given state $x$. These terminal sets sacrifice the theoretical guarantee of persistent feasibility for the FHOCP in favor of reduced computational complexity for fast online control. In particular, for a given $x$, we take the $K$-nearest neighbors from each of the previous $P$ laps by evaluating the weighted Euclidean distance over the relevant states:
\begin{align} \label{eq:knn_distance}
    (x_k^i - x)^\top D (x_k^i - x), \ \forall &i \in \{j, j-1, \dots, j-(P-1)\}, \nonumber\\
    &k \in \{0, 1, \dots, T^i\}
\end{align}
for some given $P$ and $D \succeq 0$. Letting $\mathbf{X}^j(x;\mathcal{D}^j) \in \mathbb{R}^{n\times KP}$ denote the matrix formed by the $KP$ states closest to $x$, we then define the target set as the convex hull of these states: $\mathcal{X}_N^j(x;\mathcal{D}^j) = \{\bar{x} \in \mathbb{R}^n \ | \ \exists \lambda \in \mathbb{R}^{KP}, \ 0 \leq \lambda \leq 1, \ \mathbf{1}^{\top}\lambda = 1, \mathbf{X}^j(x;\mathcal{D}^j) \lambda = \bar{x}\}$.
% \begin{align*}
%     \mathcal{X}_N^j(x;\mathcal{D}^j) = \{\bar{x} \in \mathbb{R}^n \ | \ \exists \lambda \in \mathbb{R}^{KP}&, \ 0 \leq \lambda \leq 1, \ \mathbf{1}^{\top}\lambda = 1, \nonumber\\ 
%     &\mathbf{X}^j(x;\mathcal{D}^j) \lambda = \bar{x}\}
% \end{align*}

\subsection{Local Terminal Cost-to-go Function}

By assumption, the closed-loop state trajectories stored in $\mathcal{D}^j$ reach the finish line in a finite amount of time. We can therefore compute the cost-to-go for any state $x_k^i$ in $\mathcal{D}^j$ as $T^i - k$, which represents the time remaining to finish the $i$-th lap for $i \in \{0, \dots, j\}$. Let $J_N^j(x;\mathcal{D}^j) \in \mathbb{R}^{KP}$ denote the vector of cost-to-go values calculated for the states in $\mathbf{X}^j(x;\mathcal{D}^j)$. We can therefore define the minimum cost-to-go function for any $\bar{x} \in \mathcal{X}_N^j(x;\mathcal{D}^j)$ as:
\begin{align*}
    Q_N^j(\bar{x},x;\mathcal{D}^j) \\
    = \min_{\lambda \in \mathbb{R}^{KP}} \ & \ J_N^j(x;\mathcal{D}^j)^\top \lambda \\
    \text{subject to} \ & \ 0 \leq \lambda \leq 1, \ \mathbf{1}^{\top}\lambda = 1, \ \mathbf{X}^j(x;\mathcal{D}^j)\lambda = \bar{x}
\end{align*}
This function can be thought of as a local convex approximation to the optimal cost-to-go (from solving \eqref{eq:inf_horizon_min_lap_prob}) at iteration $j$, which when incorporated as the terminal cost-to-go function, allows an FHOCP to reason about infinite-horizon behavior of the vehicle, to the extent that it is captured in the dataset.

\subsection{Local LMPC}

Using the terminal target set and cost-to-go function synthesized from the dataset $\mathcal{D}^{j-1}$, we may now construct the FHOCP at time step $k$ of iteration $j$ for our local LMPC policy as follows:
\begin{subequations} \label{eq:local_lmpc}
    \begin{align}
        J_k^j(x_k,u&_{k-1},\bar{\mathbf{z}}_k;\mathcal{D}^{j-1}) = \nonumber\\
        \min_{\mathbf{x}, \mathbf{u}, \lambda} \ & \ \sum_{t=0}^{N-1}\mathds{1}_{\mathcal{F}}(x_t) + c_u \|u_t\|_2^2 + c_{\Delta u} \|u_t-u_{t-1}\|_2^2 \nonumber \\
        & \qquad + J_N^{j-1}(\bar{x}_{k+N};\mathcal{D}^{j-1})^\top \lambda \label{eq:lmpc_cost} \\
        \text{subject to} \ & \ x_0 = x_k, \ u_{-1} = u_{k-1}\\
        & \ x_{t+1} = A(\bar{z}_{k+t};\mathcal{D}^{j-1})x_t + B(\bar{z}_{k+t};\mathcal{D}^{j-1})u_t \nonumber\\ 
        & \quad + C(\bar{z}_{k+t};\mathcal{D}^{j-1}), \ t \in \{0, \dots, N-1\}, \label{eq:local_lmpc_dynamics} \\
        & \ x_t \in \mathcal{X}, \ u_t \in \mathcal{U}, \ t \in \{0, \dots, N\} \\ 
        & \ \mathbf{X}^{j-1}(\bar{x}_{k+N};\mathcal{D}^{j-1})\lambda = x_N, \\
        & \ 0 \leq \lambda \leq 1, \ \mathbf{1}^{\top}\lambda = 1
    \end{align}
\end{subequations}
where $\bar{z}_k = (\bar{x}_k, \bar{u}_k)$ and  $\bar{\mathbf{z}}_k = \{\bar{z}_k, \dots, \bar{z}_{k+N}\}$ is a state and input sequence which is used to form the local convex approximation. In practice, this is chosen as the solution to \eqref{eq:local_lmpc} from the previous time-step. Construction of the $A$, $B$, and $C$ matrices in the ATV dynamics \eqref{eq:local_lmpc_dynamics} is the main contribution of this work and will be discussed in the next section.
% We also propose the adaptive parameter $\alpha_k\in[0,1]$ in \eqref{eq:lmpc_cost}, which determines the relative importance of minimizing the cost-to-go function. Whereas this parameter is set to a constant value of $\alpha_k = 1$ in \cite{rosolia_learning_2017}, we propose a simple approach to adapt the value of the parameter based on properties of the dataset used for synthesis of the ATV dynamics matrices. 
Note that \eqref{eq:local_lmpc} is a convex program which can be solved efficiently using existing solvers.

Finally, let $\mathbf{u}^\star = \{u_0^\star, \dots, u_{N-1}^\star\}$ be the optimal control sequence which solves \eqref{eq:local_lmpc} at time step $k$. The input $u_k = u_0^\star$ is applied to the vehicle and the FHOCP \eqref{eq:local_lmpc} is repeated at time step $k+1$ for the measured state $x_{k+1}$ until the vehicle reaches the finish line, at which point $\mathcal{D}^j$ will be constructed using the closed-loop trajectory for lap $j$. 

\section{Learned Vehicle Dynamics Model} \label{sec:learned_model}

We now introduce the main contribution of our work, which is a learning strategy on the dynamics states $v_x,v_y,\omega_z$ based on regressive error. The learning process takes as inputs the nominal dynamics model and a set of neighboring states, and outputs a learned local affine approximation of the true vehicle dynamics. 
We assume that the true dynamics of the system can be written as follows
\begin{align} \label{eq:nonlinear_system_and_error}
    x^+ = f(x, u) + e(x, u),
\end{align}
where $f$ are the nominal dynamics from \eqref{eq:nominal_dynamics_model} and $e$ denotes some unknown modeling error which we would like to identify using data. For a given state and input $x$ and $u$, let us denote the prediction of the nominal model as $\hat{x}^+ = f(x, u)$. We may now linearize the error dynamics about a reference state and input $\bar{z}$ as follows:
\begin{align} \label{eq:regression_system}
    x^+ - \hat{x}^+ = A^e x + B^e u + C^e,
\end{align}
where $A^e$ and $B^e$ are the Jacobians of $e$ w.r.t. $x$ and $u$ evaluated at $\bar{z}$ and $C^e = e(\bar{x},\bar{u})-A^e\bar{x}-B^e\bar{u}$. It can be clearly seen from \eqref{eq:regression_system} that the linearization of the unknown error term is related to the system state and input $x$ and $u$, the actual state evolution $x^+$, and the prediction of the nominal model $\hat{x}^+$, which are all quantities that can be easily obtained from the dataset $\mathcal{D}^{j-1}$ at iteration $j$. As such, we query $\mathcal{D}^{j-1}$ to find the $M$ nearest neighbors $\mathbf{z} = \{z_1, \dots ,z_{M}\}$ of $\bar{z}$ and their corresponding state evolutions $\mathbf{x}^+ = \{x_1^+,\dots ,x_{M}^+\}$ (using a technique similar to \eqref{eq:knn_distance}). We then compute the prediction of the nominal model at each of the state and input pairs in $\mathbf{z}$ to obtain $\hat{\mathbf{x}}^+ = \{\hat{x}_1^+,\dots ,\hat{x}_M^+\}$. As mentioned before, we are interested in learning only the error dynamics associated with the velocity components of the state $x$, as the kinematic components of the state are well-understood. Therefore, using the datasets $\mathbf{z}$, $\mathbf{x}^+$, and $\hat{\mathbf{x}}^+$, we define the residuals for each velocity state as follows:
\begin{align} 
    {A^{e}}[11:13] x_{m}[1:3] + {B^e}[11] a_m + {C^e}[1] &= v_{x,m}^+ - \hat{v}_{x,m}^+, \nonumber\\
    {A^e}[21:23] x_{m}[1:3] + {B^e}[22] \delta_m + {C^e}[2] &= v_{y,m}^+ - \hat{v}_{y,m}^+, \nonumber\\
    {A^e}[31:33] x_{m}[1:3] + {B^e}[32] \delta_m + {C^e}[3] &= \omega_{z,m}^+ - \hat{\omega}_{z,m}^+, \label{eq:residuals}
\end{align}
where $m\in\{1, \dots, M\}$ and we use the notation $A[11:13]$ to index the first three elements of the first row of matrix $A$. Note that we have injected additional structure into the individual residuals by making explicit the dependence of each velocity state on a certain subset of the input components. From these three expressions, we define the following regressor vectors:
\begin{align*}
    \Gamma_{v_x} = 
    \begin{bmatrix}
        {\scriptstyle {A^e}[11:13]} \\ {\scriptstyle {B^e}[11]} \\ {\scriptstyle {C^e}[1]}
    \end{bmatrix} ,
    \Gamma_{v_y} = 
    \begin{bmatrix}
        {\scriptstyle {A^e}[21:23]} \\ {\scriptstyle {B^e}[22]} \\ {\scriptstyle {C^e}[2]}
    \end{bmatrix} ,
    \Gamma_{\omega_z} = 
    \begin{bmatrix}
        {\scriptstyle {A^e}[31:33]} \\ {\scriptstyle {B^e}[32]} \\ {\scriptstyle {C^e}[3]}
    \end{bmatrix}.
\end{align*}

We solve the regression problem in a manner similar to \cite{rosolia_learning_2020}, where we weigh the importance of data point $m$ using the Epanechnikov kernel function \cite{epanechnikov_non-parametric_1969} with bandwidth $h$:
\begin{align*}
    K(u)&=\begin{cases}
        \frac{3}{4}(1-u^2/h^2), & |u| < h \\
        0, & \text{otherwise}
    \end{cases}.
\end{align*} 
such that larger penalties are assigned to the residuals for data points that are close to the linearization point $\bar{x}$.
For $l = \{v_x, v_y, \omega_z\}$, the weighted least squares problem is defined as follows:
\begin{align}
    \Gamma_l^\star = \operatorname*{argmin}_{\Gamma_l} \sum_{m=1}^M K\left(||\bar{z}-z_m||^2_Q\right)y_m^l(\Gamma_l) + \epsilon \|\Gamma_l\|_2^2, \label{eq:regression_least_squares}
\end{align}
where $Q \succeq 0$ denotes the relative scaling between the variables, $y_m^{l}(\Gamma_l)$ are the $l^2$-norms of the residuals in \eqref{eq:residuals}, and $\epsilon > 0$ is a regularization parameter.

% Finally, we can unpack the optimal regressor vectors $\Gamma_l^\star$ to construct the matrices $A^e$, $B^e$, and $C^e$, and add them to the nominal ATV dynamics to obtain the complete ATV model for \eqref{eq:local_lmpc_dynamics}:
% \begin{align*}
%     A(\bar{z};\mathcal{D}^{j-1}) &= A^f + 
%     \begin{bmatrix}
%     \begin{matrix}
%         \Gamma_{v_x}^\star[1:3] \\
%         \Gamma_{v_y}^\star[1:3] \\
%         \Gamma_{\omega_z}^\star[1:3] \\
%         \mathbf{0}_{3 \times 3}
%     \end{matrix}
%     & \mathbf{0}_{6\times 3}
%     \end{bmatrix},
%     \\
%     B(\bar{z};\mathcal{D}^{j-1}) &= B^f +
%     \begin{bmatrix}
%     \begin{matrix}
%          \Gamma_{v_x}^\star[4] & 0 \\
%         0 & \Gamma_{v_y}^\star[4] \\
%         0 & \Gamma_{\omega_z}^\star[4]
%     \end{matrix} \\
%     \mathbf{0}_{3 \times 2}
%     \end{bmatrix},
%     \\
%     C(\bar{z};\mathcal{D}^{j-1}) &= C^f + 
%     \begin{bmatrix}
%         \Gamma_{v_x}^\star[5] & \Gamma_{v_y}^\star[5] & \Gamma_{\omega_z}^\star[5] & \mathbf{0}_{1 \times 3}
%     \end{bmatrix}^\top,
% \end{align*}
% where $A^f$ and $B^f$ are the Jacobians of $f$ w.r.t. $x$ and $u$ evaluated at $\bar{z}$ and $C^f = f(\bar{x},\bar{u})-A^f\bar{x}-B^f\bar{u}$. We note that the key difference between this procedure and that in \cite{rosolia_learning_2020} is that the regressive learning is performed on the \textit{error} between the prediction of the nominal model and the actual state evolution data instead of on the entire model in \eqref{eq:nonlinear_system_and_error}. 

Finally, we can construct the matrices $A^e$, $B^e$, and $C^e$ from $\Gamma_l^\star$ as follows:
\begin{align*}
    A^e &= 
    \begin{bmatrix}
    \begin{matrix}
        \Gamma_{v_x}^\star[1:3] \\
        \Gamma_{v_y}^\star[1:3] \\
        \Gamma_{\omega_z}^\star[1:3] \\
        \mathbf{0}_{3 \times 3}
    \end{matrix}
    & \mathbf{0}_{6\times 3}
    \end{bmatrix}, \
    B^e =
    \begin{bmatrix}
    \begin{matrix}
         \Gamma_{v_x}^\star[4] & 0 \\
        0 & \Gamma_{v_y}^\star[4] \\
        0 & \Gamma_{\omega_z}^\star[4]
    \end{matrix} \\
    \mathbf{0}_{3 \times 2}
    \end{bmatrix},
    \\
    C^e &= 
    \begin{bmatrix}
        \Gamma_{v_x}^\star[5] & \Gamma_{v_y}^\star[5] & \Gamma_{\omega_z}^\star[5] & \mathbf{0}_{1 \times 3}
    \end{bmatrix}^\top.
\end{align*}
These are then added to the nominal ATV dynamics to obtain the complete ATV model for \eqref{eq:local_lmpc_dynamics}: $A(\bar{z};\mathcal{D}^{j-1}) = A^f + A^e$, $B(\bar{z};\mathcal{D}^{j-1}) = B^f + B^e$, and $C(\bar{z};\mathcal{D}^{j-1}) = C^f + C^e$, where $A^f$ and $B^f$ are the Jacobians of $f$ w.r.t. $x$ and $u$ evaluated at $\bar{z}$ and $C^f = f(\bar{x},\bar{u})-A^f\bar{x}-B^f\bar{u}$. We note that the key difference between this procedure and that in \cite{rosolia_learning_2020} is that the regressive learning is performed on the \textit{error} between the prediction of the nominal model and the actual state evolution data instead of on the entire model in \eqref{eq:nonlinear_system_and_error}. 

To understand why our approach to model learning is more robust to scarce data, consider the case where all $\|\bar{z}-z_m\|_Q$ in \eqref{eq:regression_least_squares} are close to or greater than the kernel bandwidth $h$ (i.e., when the data $z_m$ are far from the linearization point $\bar{z}$ w.r.t. $h$). When this occurs, $K(\|\bar{z}-z_m\|_Q^2) \approx 0$, which means that the regularization term in \eqref{eq:regression_least_squares} dominates the least squares objective and we can therefore conclude that the optimal solution $\Gamma_l^\star \approx 0$. When regressing the nominal dynamics matrices directly, as in \cite{rosolia_learning_2020}, this would result in setting elements of those matrices to approximately zero, which would be disastrous in terms of model accuracy. On the other hand, when regressing the error matrices, setting elements to zero would simply correspond to \emph{not} applying any data-based correction to the nominal linearized dynamics. This allows for a natural trade-off between the nominal and data-corrected models based on the quality of the dataset. Of course, this could be mitigated by the selection of a large enough bandwidth $h$. However, as this is a tuning parameter which is meant to act as a threshold on a weighted distance metric in high dimensions, it may not be immediately apparent what constitutes a good setting. We therefore want the LMPC policy to be robust to values of $h$ to allow for safe tuning of the controller.
% This strategy not only provides a safety fallback when the regressive learning fails at run-time, but also demonstrates, in comparison to \cite{rosolia_learning_2020}, improved robustness to parameter tuning and reduced modeling errors in scenarios when data is scarce.

% Finally, since $K(\|\bar{z}-z_m\|_Q^2)$ provides a scalar measure of the quality of an individual data point $z_m$, we propose to use this as a trust parameter in order to adapt the weight $\alpha_k$ in \eqref{eq:lmpc_cost}. In particular, we set $\alpha_k = (4/3)\max_{m \in \{1, \dots, M\}} K(\|\bar{z}-z_m\|_Q^2)$. 

\section{Experiments \& Results}

We now present a set of experiments in simulation and hardware to demonstrate the effectiveness of combining LMPC and error dynamics regression in autonomous racing. The key metrics throughout these experiments are

\begin{itemize}
    \item 20th-iteration lap time (ILT-20): the average lap time after running the LMPC for 20 iterations. 
    % The shorter the lap time, the better the performance of the control policy.
    \item Iteration to fail (ITF): the average number of iterations before an experiment fails due to loss of control or track boundary violation. 
    % We set the maximum number of iterations in an experiment to 20 laps around the track.
\end{itemize}

\subsection{Robustness Study on Learning Parameters}

\begin{figure}[t]
     \centering
     {\includegraphics[trim={0 0 0 0cm},clip,width=1.0\columnwidth]{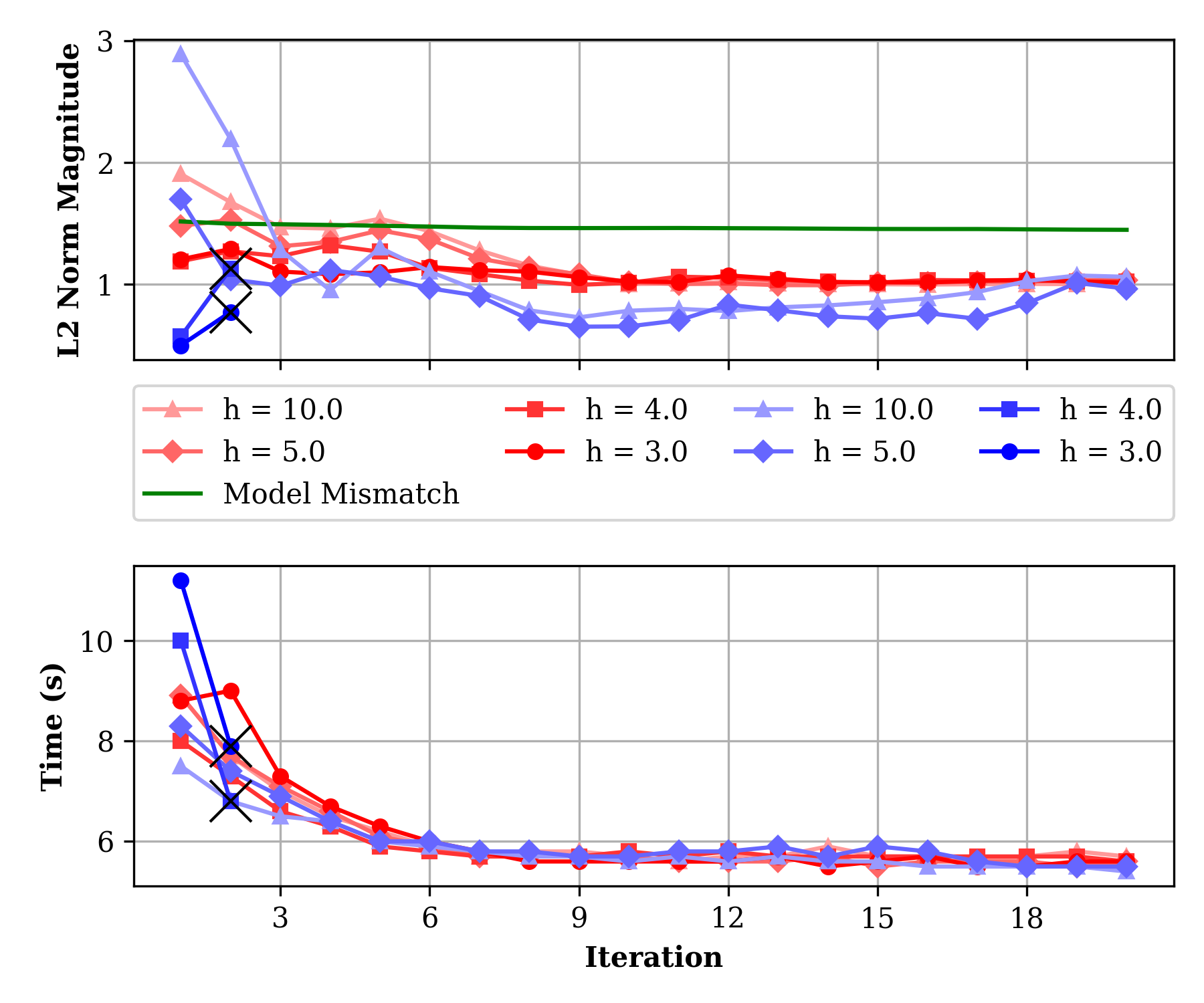}}
     \vspace{-0.8cm}
     \caption{Average error of the learned model (Top) and lap times (bottom) over iterations for various settings of the bandwidth $h$. Red and blue lines correspond to the error regression and full regression cases, respectively. A cross is placed where failure occurs due to the vehicle leaving the track.}
     \label{fig:sensitivity_a_norm}
     \vspace{-0.7cm}
\end{figure}

\begin{figure*}[t]
     \centering
     {\includegraphics[trim={0 0 0 1cm},clip,scale=0.6]{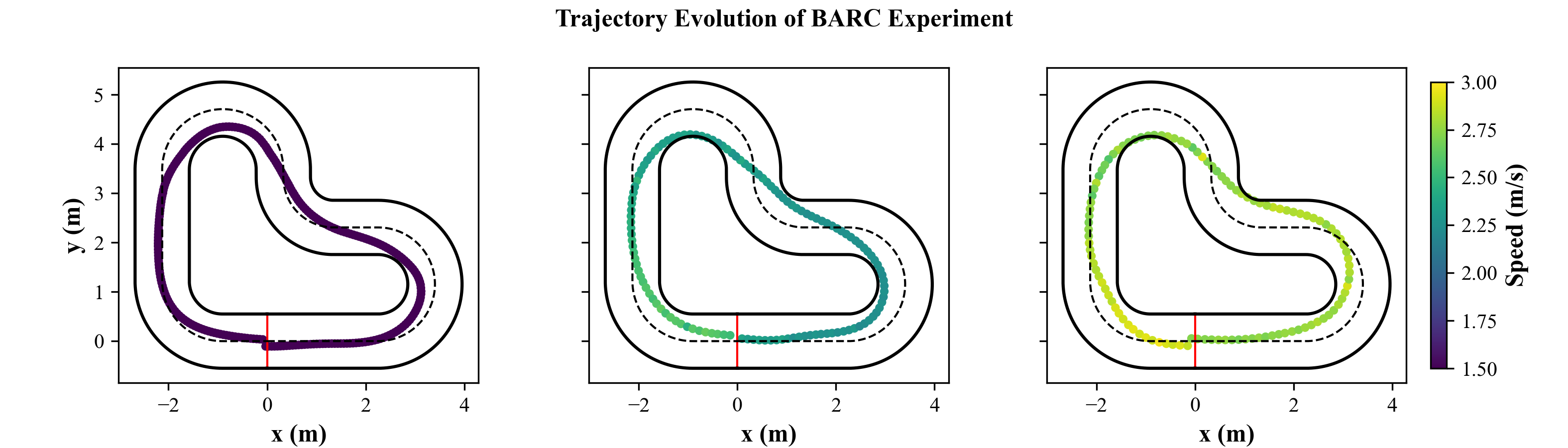}}
     \vspace{-0.2cm}
     \caption{Visualization of the 1st, 5th and 20th iteration's trajectory of LMPC with error dynamics regression on the BARC platform.}
     \label{fig:barc_vis}
     \vspace{-0.7cm}
\end{figure*}

This experiment aims to study how LMPC parameter tunings affect the safety of the learning process. The motivation is that there are not many trial-and-error opportunities when tuning parameters on safety-critical hardware such as a race car, and a robust control setup should be able to handle a broad range of parameter settings without destabilizing the system. We set up simulation experiments to compare our approach against the LMPC implementation in \cite{rosolia_learning_2020} in terms of their robustness to the change in the bandwidth parameter $h$ and the control-rate cost (CRC) ($c_{\Delta u}$ in \eqref{eq:lmpc_cost}). Intuitively, the bandwidth $h$ governs a trade off between the quality and quantity of the data used for model regression. Whereas a high bandwidth allows for potentially more data points to be selected, they can be far (in the sense of the norm $\|\cdot\|_Q$) from the linearization point. On the other hand, a low bandwidth requires that the data points be close to the linearization point, but could result in few data points being selected when data is sparse. As for the CRC, it is analogous to the ``learning rate" for LMPC. A higher CRC will penalize drastic changes in the control input and keep the vehicle closer to the safe set over the MPC horizon. Both $h$ and the CRC are therefore key parameters when tuning the behavior of the LMPC policy. For all simulation studies, we use the L-shaped track shown in Figure~\ref{fig:barc_vis} and model the rigid body with the dynamic bicycle model. The tire forces are modeled with the Pacejka tire model \cite{pacejka1992magic} with tire-road friction coefficient $\mu=0.9$.

In our first simulation experiment, we set $\text{CRC} = 0.1$ and compare the effect of the bandwidth $h$ on the accuracy of the learned model for our error regression LMPC and the full regression LMPC from \cite{rosolia_learning_2020}. In the error regression case, we use the kinematic bicycle model as the nominal model in \eqref{eq:nominal_dynamics_model} which induces mismatch with the simulation model. 
% This would result in dangerous over-confidence in cornering behavior at high speeds if no error learning is performed. 
To measure accuracy of the learned model, we compute the Frobenius norm of the difference between the learned dynamics matrix $A(\bar{z},\mathcal{D}^{j-1})$ and the linearized dynamics of the simulation model at each time step and record the average over each iteration. The results for $h = \{3, 4, 5, 10\}$ are shown in Figure~\ref{fig:sensitivity_a_norm}, where we may immediately observe that for high settings of $h$, both the error and full regression cases perform similarly in both the learned model accuracy and lap times. Importantly, we see that the error regression case can correct for the mismatch between the nominal and simulation models, which is shown by the green line. For low settings of $h$, it can be seen that while \cite{rosolia_learning_2020} fails after the second iteration, the error regression LMPC is able to remain stable and converge to similar performance as the high bandwidth cases. This is due to the reasons discussed in Section~\ref{sec:learned_model} where sparsity w.r.t. $h$ can result in significant model inaccuracies in \cite{rosolia_learning_2020}, but would naturally induce a fall back to the nominal model in the error regression case. Sparsity in the regression data is especially acute in the initial laps when the LMPC policy is rapidly improving vehicle performance as there can be significant differences in the data between successive laps. We note that the behavior of the error regression case is largely dependent on the accuracy of the nominal model when data is sparse and stability cannot be guaranteed for arbitrary realizations of the nominal model. However, especially in car racing, it is not unreasonable that to assume access to a fairly accurate nominal model which can be obtained through standard system identification approaches. Future work will examine the effect of distance between the nominal and true system on the stability of the error regression LMPC.

% To understand how the accuracy of the learned model compares between the two approaches, we inspect the average norm of the difference between the learned dynamics matrix $A(\bar{z},\mathcal{D}^{j-1})$ and the actual local dynamics matrix of the simulation model, which is computed by taking the Jacobian of the true system model at the same points $\bar{z}$. 
% Figure \ref{fig:sensitivity_a_norm} shows the evolution of the mismatch in the dynamics matrix over laps of a simulation experiment with a CRC value of 1. We highlight that although the modeling error of full dynamics regression in \cite{rosolia_learning_2020} eventually reaches the same level as our error dynamics regression, our learning approach better approximates the actual local dynamics consistently and always only shows a small model mismatch after learning. This is especially important in the initial iterations, when there is relatively scarce data to accurately fit an ATV model. We believe that this large initial model mismatch with \cite{rosolia_learning_2020} may have contributed to the failure cases with lower settings of CRC, as the LMPC is unintentionally reaching the handling limit due to exploiting a poorly fitted model before sufficient data can be collected to perform an accurate regression of the full dynamics.

\begin{figure*}[t]
     \centering
     {\includegraphics[scale=0.61]{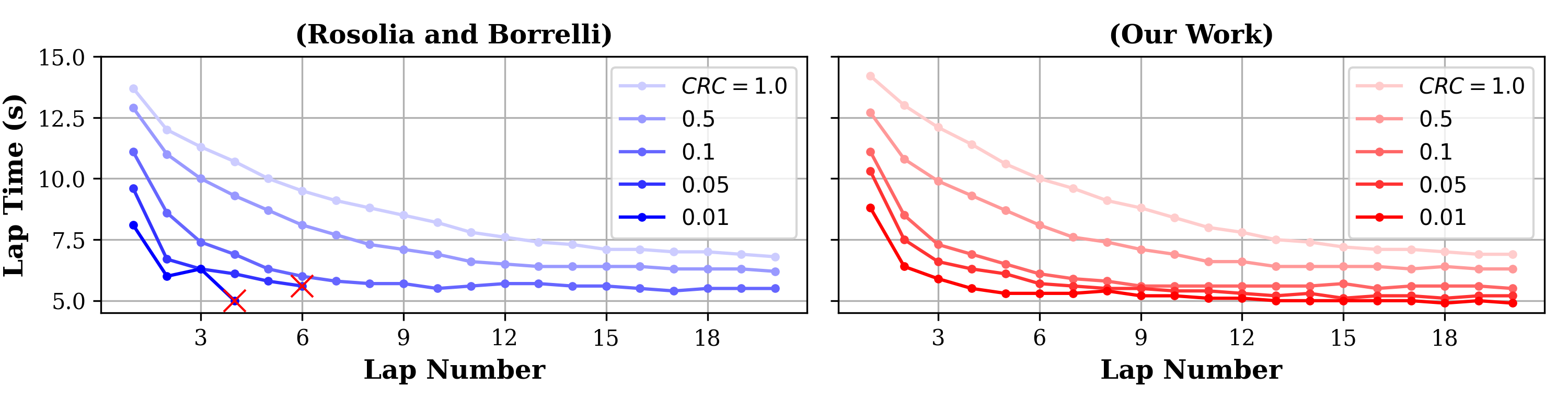}}
     \vspace{-0.4cm}
     \caption{Results from the CRC robustness study. The left and right plots show the iteration lap times over different CRC tunings for \cite{rosolia_learning_2020} and our error regression LMPC respectively. A red cross is placed where failure occurs due to the vehicle leaving the track.}
     \label{fig:sensitivity_result}
     \vspace{-0.4cm}
\end{figure*}

In our second simulation experiment, we fix the bandwidth to $h=5$ and ran both LMPC implementations with $\text{CRC}=\{1.0, 0.5, 0.1, 0.05, 0.01\}$ and recorded the lap time at every iteration. 
Like the previous study, we purposely introduce mismatch in the nominal model. Though for this study, we do so through the tire-road friction coefficient. Whereas a value of 0.9 is used in the simulation model, a value of 1.2 is chosen for the LMPC nominal dynamics model in \eqref{eq:nominal_dynamics_model}, which is typically only seen on full-scale race cars with racing slick tires. This would have induced dangerous over-confidence in the model about the surface friction, as shown in the following hardware experiments, if no dynamics learning is used.
Table \ref{tab:sensitivity_result} presents the key results of this experiment, 
% When sweeping the CRC over the values of  $\{1.0, 0.5, 0.1, 0.05, 0.01\}$, 
where we observe that with CRC penalties of 1.0, 0.5, and 0.1, both \cite{rosolia_learning_2020} and our approach show similar performance at the end of 20 laps, with \cite{rosolia_learning_2020} achieving slightly faster lap time reduction as evidenced in Figure~\ref{fig:sensitivity_result}. However, once we decrease the CRC below 0.1, we see that \cite{rosolia_learning_2020} fails to complete the 20 laps, whereas our method exhibits greater robustness to different settings of the CRC and can still converge to high performing lap times within 20 iterations. 
% Figure \ref{fig:sensitivity_result} compares the per-iteration performance of both methods, showing that \cite{rosolia_learning_2020} fails with more than half of the CRC settings, while our approach was able to keep the system safe for all settings.

Overall, our simulation experiments suggest that LMPC with error dynamics regression exhibits greater stability and robustness to key parameter settings and works well even when scarce dynamics data are available. These properties could potentially make the system more suitable for hardware deployment in safety-critical applications.

\begin{table}[t]
\centering
\begin{tabular}{c|cc|cc}
\toprule
\multirow{2}{*}{\textbf{Control Rate Cost}} & \multicolumn{2}{c|}{\textbf{ILT-20}} & \multicolumn{2}{c}{\textbf{ITF}} \\ \cline{2-5} 
 & \multicolumn{1}{l|}{\cite{rosolia_learning_2020}} & \multicolumn{1}{l|}{ours} & \multicolumn{1}{l|}{\cite{rosolia_learning_2020}} & \multicolumn{1}{l}{ours} \\ \midrule
1.0 & \multicolumn{1}{c|}{\textbf{6.4}} & 6.5 & \multicolumn{1}{c|}{20+} & 20+ \\
0.5 & \multicolumn{1}{c|}{\textbf{6.0}} & 6.2 & \multicolumn{1}{c|}{20+} & 20+ \\
0.1 & \multicolumn{1}{c|}{\textbf{5.5}} & 5.6 & \multicolumn{1}{c|}{20+} & 20+ \\
0.05 & \multicolumn{1}{c|}{-} & \textbf{5.2} & \multicolumn{1}{c|}{6} & \textbf{20+} \\
0.01 & \multicolumn{1}{c|}{-} & \textbf{5.0} & \multicolumn{1}{c|}{4} & \textbf{20+} \\ \bottomrule
\end{tabular}
\caption{Results of the simulation robustness study. A hyphen in the 20th-Iteration Lap Time (ILT-20) means that the algorithm fails before the 20th iteration. An Iteration to Fail (ITF) of 20+ means that the algorithm did not fail in the 20-iteration experiment.}
\label{tab:sensitivity_result}
\vspace{-0.2cm}
\end{table}

\subsection{Hardware Experiment Comparing Learning Performance}

\begin{figure*}[t]
     \centering
     {\includegraphics[scale=0.62]{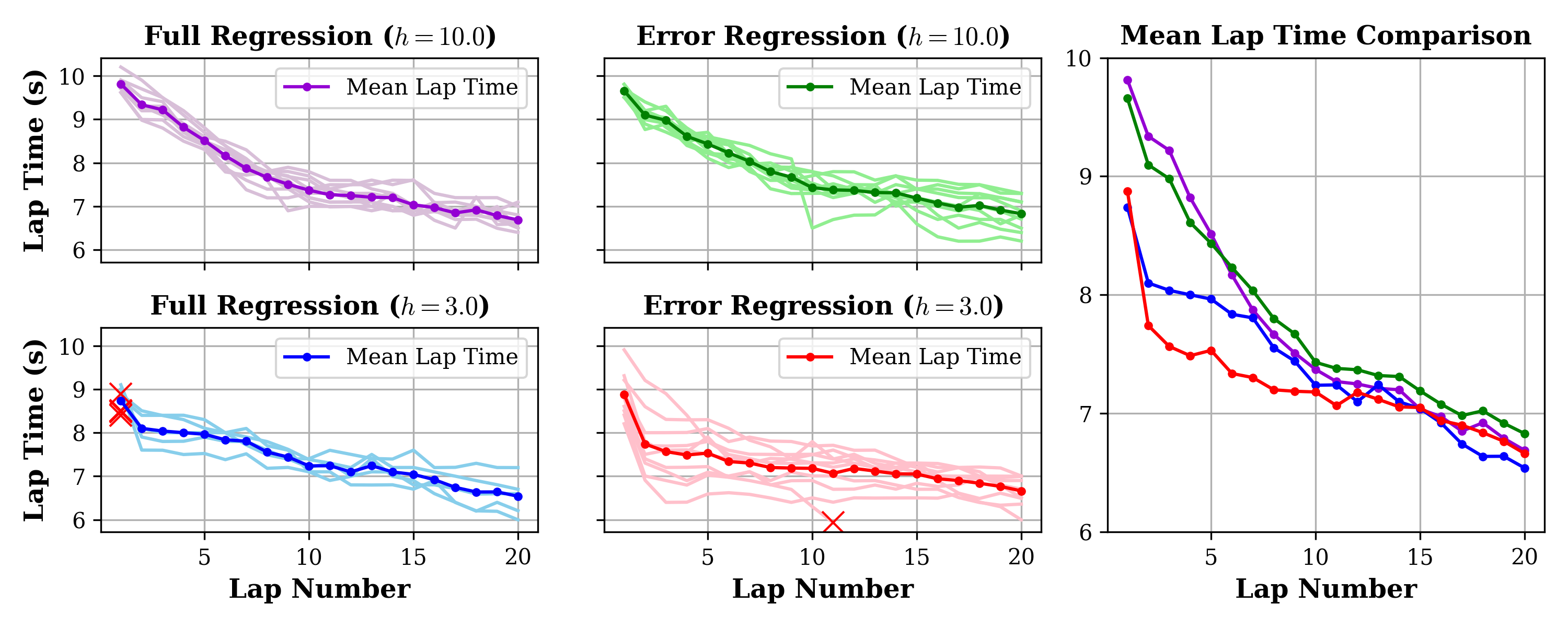}}
     \vspace{-0.3cm}
     \caption{Per-iteration results of the BARC hardware experiment. The four smaller plots on the left show the lap-time reduction of the 10 trials. The combined plot on the right compares the average lap times achieved. A red cross is placed where failure occurs due to the vehicle leaving the track.}
     \label{fig:barc_result}
     \vspace{-0.3cm}
\end{figure*}

\begin{table}[t]
\centering
\begin{tabular}{l|c|c}
\toprule
\textbf{LMPC Configuration} & \textbf{Avg. ILT-20} & \textbf{Avg. ITF} \\ \midrule
Full Regression (\cite{rosolia_learning_2020}, $h=10.0$) & 6.69 & 20+ \\
Error Regression (ours, $h=10.0$) & 6.82 & 20+ \\
Full Regression (\cite{rosolia_learning_2020}, $h=3.0$) & 6.53 & 10.5 \\
Error Regression (ours, $h=3.0$) & 6.66 & 19.1 \\ \bottomrule
\end{tabular}
\caption{Results of the BARC hardware experiment.}
\label{tab:barc_result}
\vspace{-0.7cm}
\end{table}

In this section, we present results from physical experiments with the Berkeley Autonomous Race Car (BARC) platform. These vehicles are based on 1/10th scale, off-the-shelf RC cars that have been modified for autonomous racing. LMPC computation is done on a laptop computer with a 2.6 GHz 9th-Gen Intel Core i7 CPU, localization is provided by a motion capture system, and communication with the vehicle is handled with ROS2. We conducted two sets of experiments to demonstrate the effect of the bandwidth $h$ on the stability and learning performance of \cite{rosolia_learning_2020} and our error regression LMPC. Each set of experiments consists of 10 runs of at most 20 laps with $\text{CRC}=0.1$. We record the lap times and record any failure cases before the 20-lap threshold.

% \begin{table}[t]
% \centering
% \begin{tabular}{l|l|l}
% \toprule
% \textbf{Specification}        & \textbf{\textbf{BARC}}                                      & \textbf{\textbf{IAC Car}}                                  \\ \midrule
% Mass                          & 2.2 \si{\kilo\gram}                                         & 800 \si{\kilo\gram}                                        \\
% Wheelbase                     & 0.26  \si{\meter}                                           & 4  \si{\meter}                                             \\
% Top Speed                     & $\sim$10 \si[per-mode=symbol]{\kilo\meter\per\hour}         & $\sim$320 \si[per-mode=symbol]{\kilo\meter\per\hour}       \\
% Top Lateral Acceleration & $\sim$10-15 \si[per-mode=symbol]{\meter\per\square\second}  & $\sim$15-20 \si[per-mode=symbol]{\meter\per\square\second}                                              \\
% Top Longitudinal Acceleration & $\sim$10-15  \si[per-mode=symbol]{\meter\per\square\second} & $\sim$10-15 \si[per-mode=symbol]{\meter\per\square\second} \\
% Top Braking                   & $\sim$10-15  \si[per-mode=symbol]{\meter\per\square\second} & $\sim$20-30 \si[per-mode=symbol]{\meter\per\square\second} \\
% Max Engine Power              & $\sim$100 \si{\watt}                                        & $\sim$340  \si{\kilo\watt}                                 \\ \bottomrule
% \end{tabular}
% \caption{Vehicle specifications of BARC and IAC platforms used in this work.}
% \label{tab:vehicle_specs}
% \vspace{-0.3cm}
% \end{table}

In the first experiment, we run \cite{rosolia_learning_2020} and our error regression LMPC with $h=10$. This corresponds to the high bandwidth case in the simulation study where both policies achieved similar performance and were able to successfully complete 20 laps. The results of our experiment are shown in the top row of plots in Figure~\ref{fig:barc_result}, where we again see that the two approaches show similar performance in lap time reduction and are able to remain stable over 20 laps. This corroborates our simulation study and shows that with proper tuning, our error regression LMPC is essentially equivalent to \cite{rosolia_learning_2020}. In the second experiment, we run \cite{rosolia_learning_2020} and our error regression LMPC with $h=3$. This corresponds to the low bandwidth case in the simulation study where we data sparsity has a destabilizing effect on \cite{rosolia_learning_2020}. The results of our experiment are shown in the bottom row of plots in Figure~\ref{fig:barc_result}, which also supports our observations from the simulation study. In particular, we observe that when a low bandwidth setting is used, \cite{rosolia_learning_2020} failed in 5 of the 10 trials, whereas our error learning LMPC only failed in a single trial.

\subsection{Hardware Demonstration on a Full-Size Race Car}

In this demonstration, we deployed our LMPC and error dynamics regression strategy on a full-size \textit{IndyLights} race car which has been used in the Indy Autonomous Challenge \cite{wischnewski2022indy}. The race track used for this demonstration is the Putnam Park Road Course in Indiana, USA. We construct the initial dataset $\mathcal{D}^0$ for the vehicle by running a tracking MPC on the center line of the track at 10 \si[per-mode=symbol]{\meter\per\second} for 3 laps. We note that this step could also be done with a more rudimentary controller. We then ran LMPC with error dynamics regression for 10 iterations. Figure \ref{fig:iac_vis} visualizes the trajectory driven by the vehicle, with a zoomed-in view to highlight how LMPC optimizes the shape of the trajectory for a hairpin turn. It can be clearly seen in Figure~\ref{fig:iac_vis} that the vehicle starts at lower speeds closer to the center line and incrementally increases its speed through the corners and approaches the track boundary constraint.

% \begin{figure}[t]
%      \centering
%      {\includegraphics[width=\columnwidth]{figures/fig_iac_trajectory.eps}}
%      \vspace{-1.3cm}
%      \caption{Overlay of the IAC car's trajectory running LMPC with error dynamics regression. The zoomed-in image of the hairpin turn shows iterative improvement towards optimal cornering.}
%      \label{fig:iac_vis}
% \end{figure}

\section{Conclusion}

This work proposes a new error dynamics learning approach under the LMPC framework to iteratively and safely explore the boundary of vehicle handling limit in autonomous racing. We based the learning on a nominal dynamics model, which provides explanability and safety fallback for the learning process. We then derived the regression method on top to learn a local approximation of the nonlinear model mismatch. Through simulation and hardware experiments, we showed that our method exhibits greater robustness against parameter tuning and data scarcity compared with previous LMPC works, and offers new potential solutions to address the challenge of vehicle dynamics data acquisition in high-speed autonomous racing.

\bibliographystyle{IEEEtran}
\bibliography{lmpc-paper}

\end{document}